\documentclass[10pt,twocolumn,letterpaper]{article}

\usepackage{iccv}

\usepackage{graphicx}
\usepackage{amsmath}
\usepackage{amssymb}
\usepackage{booktabs}
\usepackage{diagbox}
\usepackage{amsmath}
\usepackage{amsthm,amsmath,amssymb}
\usepackage{mathrsfs}
\usepackage{amsmath}
\usepackage{amssymb}
\usepackage[dvipsnames]{xcolor}
\usepackage{times}
\usepackage{epsfig}
\usepackage{graphicx}
\usepackage{amsmath}
\usepackage{amssymb}
\usepackage{subcaption}
\usepackage[ruled,vlined]{algorithm2e}
\usepackage{float}
\usepackage{multicol}
\usepackage{multirow}
\usepackage{booktabs}
\usepackage{soul}

\usepackage[pagebackref=true,breaklinks=true,letterpaper=true,colorlinks,bookmarks=false]{hyperref}

 \iccvfinalcopy 

\def\iccvPaperID{XXXX} 

\ificcvfinal\fi

\begin{document}

\title{Seeing through Unclear Glass: Occlusion Removal with One Shot}

\author{Qiang Li\thanks{Equal contribution}\\
McMaster University\\
Electrical and Computer Engineering\\
{\tt\small li1210@mcmaster.ca}
\and
Yuanming Cao\footnotemark[1]\\
McMaster University\\
Computer and Software Engineering\\
{\tt\small caoy15@mcmaster.ca}
}


\maketitle
\ificcvfinal\thispagestyle{empty}\fi

\begin{abstract}
Images taken through window glass are often degraded by contaminants adhered to the glass surfaces. Such contaminants cause occlusions that attenuate the incoming light and scatter stray light towards the camera.  Most of existing deep learning methods for neutralizing the effects of contaminated glasses relied on synthetic training data.  Few researchers used real degraded and clean image pairs, but they only considered removing or alleviating the effects of rain drops on glasses.
This paper is concerned with the more challenging task of learning the restoration of images taken through glasses contaminated by a wide range of occluders, including muddy water, dirt and other small foreign particles found in reality.  To facilitate the learning task we have gone to a great length to acquire real paired images with and without glass contaminants.  More importantly, we propose an all-in-one model to neutralize contaminants of different types by utilizing the one-shot test-time adaptation mechanism.  It involves a self-supervised auxiliary learning task to update the trained model for the unique occlusion type of each test image. Experimental results show that the proposed method outperforms the state-of-the-art methods quantitatively and qualitatively in cleaning realistic contaminated images, especially the unseen ones.
\end{abstract}


\section{Introduction}
\begin{figure}[t!]
    \centering
    \begin{subfigure}{0.48\linewidth}
    \includegraphics[width =\linewidth]{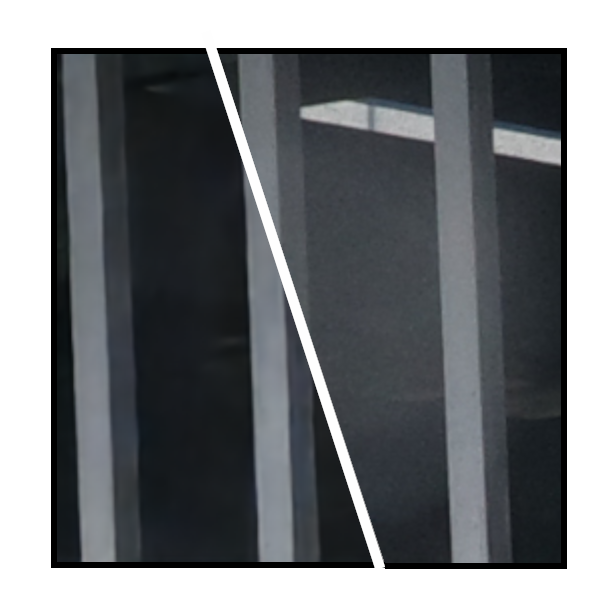}
    \caption{Dirt removal\\ \quad}
    \end{subfigure}
    \begin{subfigure}{0.48\linewidth}
    \includegraphics[width =\linewidth]{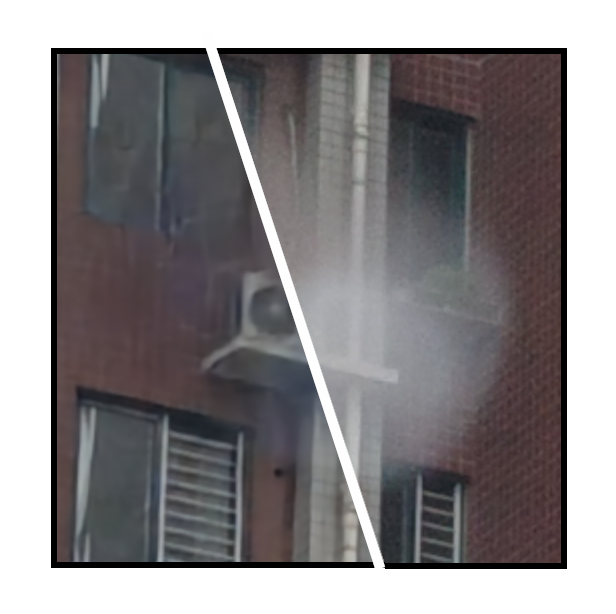}
    \caption{Raindrops removal \\ \quad }
    \end{subfigure}

    \begin{subfigure}{0.48\linewidth}
    \includegraphics[width =\linewidth]{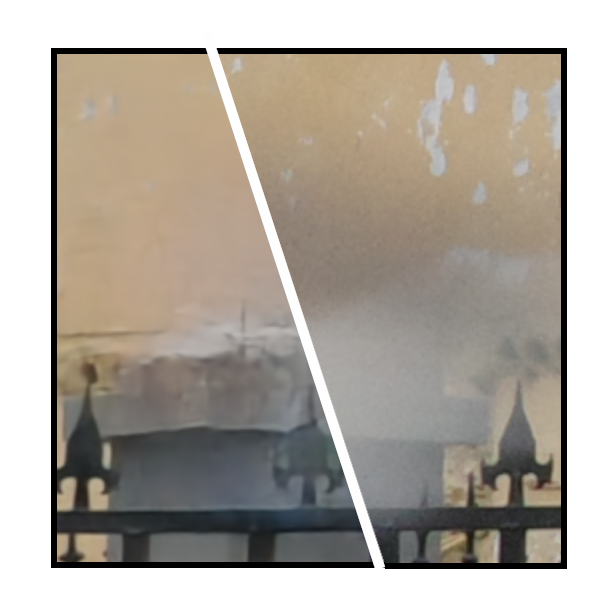}
    \caption{Muddy water removal }
    \end{subfigure}
    \begin{subfigure}{0.48\linewidth}
    \includegraphics[width =\linewidth]{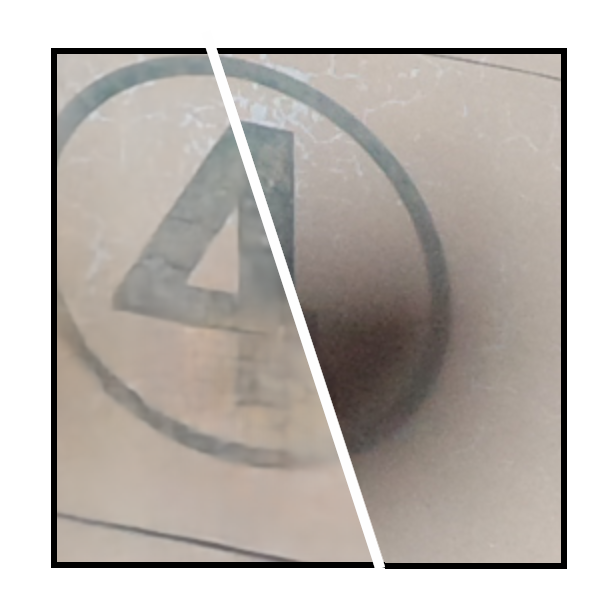}
    \caption{Particles removal }
    \end{subfigure}

    \caption{\textbf{Occlusion removal of the unclear glass.} We remove various occlusion types by the proposed all-in-one learning-based method with one shot. The occlusions contain (a) dirt, (b) raindrops, (c) muddy water, (d) particles. Our method achieves promising occlusion removal results. }
    \label{fig:compare_1}
\end{figure}


It is more often than not that we have to photograph through glasses, \eg shooting a camera inside a building or car.  In such cases, the glass, an unavoidable intermediate layer, may have small contaminants adhered to it.
When the camera is close to the glass, the contaminants create occlusion artifacts and degrade the image quality.  Such problems can easily occur with surveillance cameras covered by protection glasses that are polluted by dirt and muddy water accumulated over time, and with dash cameras in  cars when road debris and raindrops adhere to the windshield.
Solving the problem by actively cleaning the glasses or retaking the images from different angles is impractical.

With the emergence of deep learning in computer vision~\cite{dong2015image,zhang2017beyond,qian2018attentive, zhang2021dual, Xu_2025_CVPR,xu20253dgs, xu2024fast}, deep neural networks (DNNs) based approaches have appeared for specific occlusion removal, \eg dust, dirt and raindrops. These methods are suitable for images taken by consumer-level cameras.
However, they may require multiple images of the same scene with varying captures, which are inapplicable for some real-world applications\cite{liu2020learning, hirohashi2020removal, li2021let}.
Furthermore, these methods utilize a generic trained model for all testing scenarios, which is sub-optimal.  The features learned from the external training data may not generalize well on unseen occlusions where domain shift occurs~\cite{park2020fast, chi2021test}.

In this work, we first demonstrate that the occlusion removal is the same as the combination of defocus deblurring and inpainting tasks~\cite{pathak2016context}. Optical features of occlusion are obtained by investigating the image formation model and the defocus imaging model. The occlusion is decomposed into partial or complete occlusions with different characteristics. The region of partial occlusion is blurred heavily, as the irradiance is due to the defocus of both obstructions and the background scene. The region of complete occlusion is underexposed severely, as the cameras are exposed to the background properly.
Instead of synthesizing the images, 
we propose a dataset by collecting occlusion and clean image pairs. 
All images contain regions of partial occlusion, while images with thick occlusion also contain regions of complete occlusion. An advantage of utilizing real-world images is that they reduce the domain shift when testing the trained model on real images~\cite{chi2021test}.

To tackle the problem of occlusion removal task, we introduce an all-in-one framework that aims to remove various occlusion types. Instead of enlarging the lens aperture or removing occlusions by multiple shots \cite{quan2021removing, li2020all, guo2020joint, guo2018fast}, our framework requires only one shot image. Furthermore, to improve the generalization, especially on unseen occlusion types, we employ test-time adaptation mechanism to further reduce the domain shift. Specifically, we propose to attach a self-supervised auxiliary task of occlusion reconstruction alongside the primary occlusion removal task \cite{liu2018learning, jaderberg2016reinforcement, valada2018deep, zhou2017unsupervised}. Two tasks are jointly trained and share most of the weights. There are two advantages of using auxiliary task: 1) it aims to learn the optical features of occlusions, and acts as a regularization to improve the performance of the primary task.  2) during the testing phase, for each image, the model can be further updated via the auxiliary task to learn the underlying properties of partial and complete occlusion. The primary task is then tailored to remove the specific occlusion.

Our contributions can be summarized as follows:
\begin{itemize}
    \item \textbf{Occlusion Imaging Model}: 
We dive deep into an under-explored image shooting occlusion scenario where the camera is interfered by a marginally distant unclear glass. We present an unified degradation model caused by occlusions to demonstrate the necessity of occlusion removal by the neural network.

    \item \textbf{Occlusion Dataset}: We propose and collect occlusion/clean image pairs which contain various contaminants (dirt, raindrop, muddy water and particles). To the best of our knowledge, it is the first real-world occlusion removal dataset for occlusion removal with one shot (OROS). We will opensource our code and dataset as a promise.

    \item \textbf{Occlusion Removal Network}: We propose an all-in-one framework with self-supervised test-time adaptation to remove various occlusion types with one shot. For each test image, our model gets updated via self-supervision to learn the unique occlusion properties to improve the occlusion removal.

\end{itemize} 

\subsection{Methods of Optical Hardware }

Existing approaches for occlusion removal rely on increasing the imaging aperture \cite{shi2022seeing, guo2022lensfree}.
One could see through foreground occlusions by light field camera arrays and synthetic aperture refocusing technique \cite{wilburn2005high, vaish2006reconstructing, ding2024lensfree}.
Besides, Optical cloaking devices typically redirect light around the region to hide from an observer \cite{howell2014amplitude, cui2023cloud}.
In addition, the differentiable optics design in computational photography  optimize jointly by optics and algorithms in monocular depth and 3D imaging \cite{chang2019deep, wu2019phasecam3d, li2020rapid}.
While these methods can effectively suppress obstructions, they require complex optical hardware modifications.

\subsection{Methods of Image Processing}

The single image method fills the occlusion region to restore a visually pleasing result by nearby visible content and global priors.
Scott \textit{et al.} model the partial occlusions using matting, with the alpha value determined by the convolution of a kernel with a pinhole projection \cite{mccloskey2010removal}.
Besides, only a few neural networks remove raindrops in the real dataset \cite{quan2019deep, qian2018attentive}.
There currently needs to be a dataset for many kinds of occlusion removal.

The imaging systems usually utilize multiple temporal or spatial captures to identify occlusions and reconstruct background scenes.
The pioneering work removes occlusions by a simple calibration step, taking multiple images with different apertures, or relying on the statistics of natural images for post-processing videos \cite{gu2009removing}. 
Xiaoyu \textit{et al.} leverage the corresponding clean pixels from adjacent frames by a flow completion module in moving cameras \cite{li2021let}.
Yu-Lun \textit{et al.} use the motion differences between the background and the occlusions to recover both layers \cite{liu2020learning}. 
While multi-frame occlusion removal fundamentally needs many images to recover the single frame,  our method is a one-shot approach which relies on a test image.

\section{Imaging Model of Occlusions}

This section highlights the differences between our occlusion shooting scenario and others first.
Then, we acquire the OROS dataset in the proposed occlusion shooting scenario. 
Finally, we derive a unified degraded occlusion expression to reveal that occlusion removal is equivalent to both defocus deblurring and inpainting tasks.

\subsection{Differences of Shooting Scenarios}

In Fig. \ref{fig: differences}, we exhibit the different camera shooting scenarios through unclear glass (adhered by contaminants):
\noindent\textbf{Contaminants}: The unclear glass is exceedingly close to the camera lens in Fig. \ref{fig: differences} (a). The appressed unclear glass is similar to the camera lens with dirt \cite{li2021let}. 
Correspondingly, the imaging aperture becomes smaller. 

Therefore, the high-frequency components are partially preserved in the final artifact image, shot at a lower optical intensity. 

\noindent\textbf{Obstructions}: The unclear glass is close to the target as in Fig. \ref{fig: differences} (b). 
In this shooting scenario, a foreground blocks the target. 
Obstructions \cite{liu2020learning} (or scene occlusions \cite{zhan2020self})
are mainly caused by reflections (such as windows) or scenes (such as fences).
In such cases, most obstructions show clear structures and edges. Minor defocus is negligible.

\noindent\textbf{Occlusions}: The unclear glass is marginally close to the camera only as in Fig. \ref{fig: differences} (c). 
The unclear glass is severely out of focus \cite{mccloskey2010removal},
and every occlusion in such images can be decoupled into two types (as the embedded circles on the left of Fig. \ref{fig: differences} (c):
1) partial occlusion (the outer loop): similar to obstruction but is a bit blurry. Almost all  partial occlusion regions show noticeable defocus blur; 
2) complete occlusion (the inner circle): compared with contaminant images at a lower intensity, the complete occlusion shows underexposure with even opaque blur.
Therefore, the proposed occlusion shooting scenario is totally different from others (contaminants and obstructions).

This paper focuses on the occlusion scenario
by seeing through the unclear glass as in Fig. \ref{fig: differences} (c), rather than the other or general occlusions.
The occlusion-degraded images show severely defocus blur and opaque underexposure.

 \renewcommand{\thefigure}{2}
\begin{figure}[!t]
\begin{center}
    \includegraphics[width =0.8\linewidth]{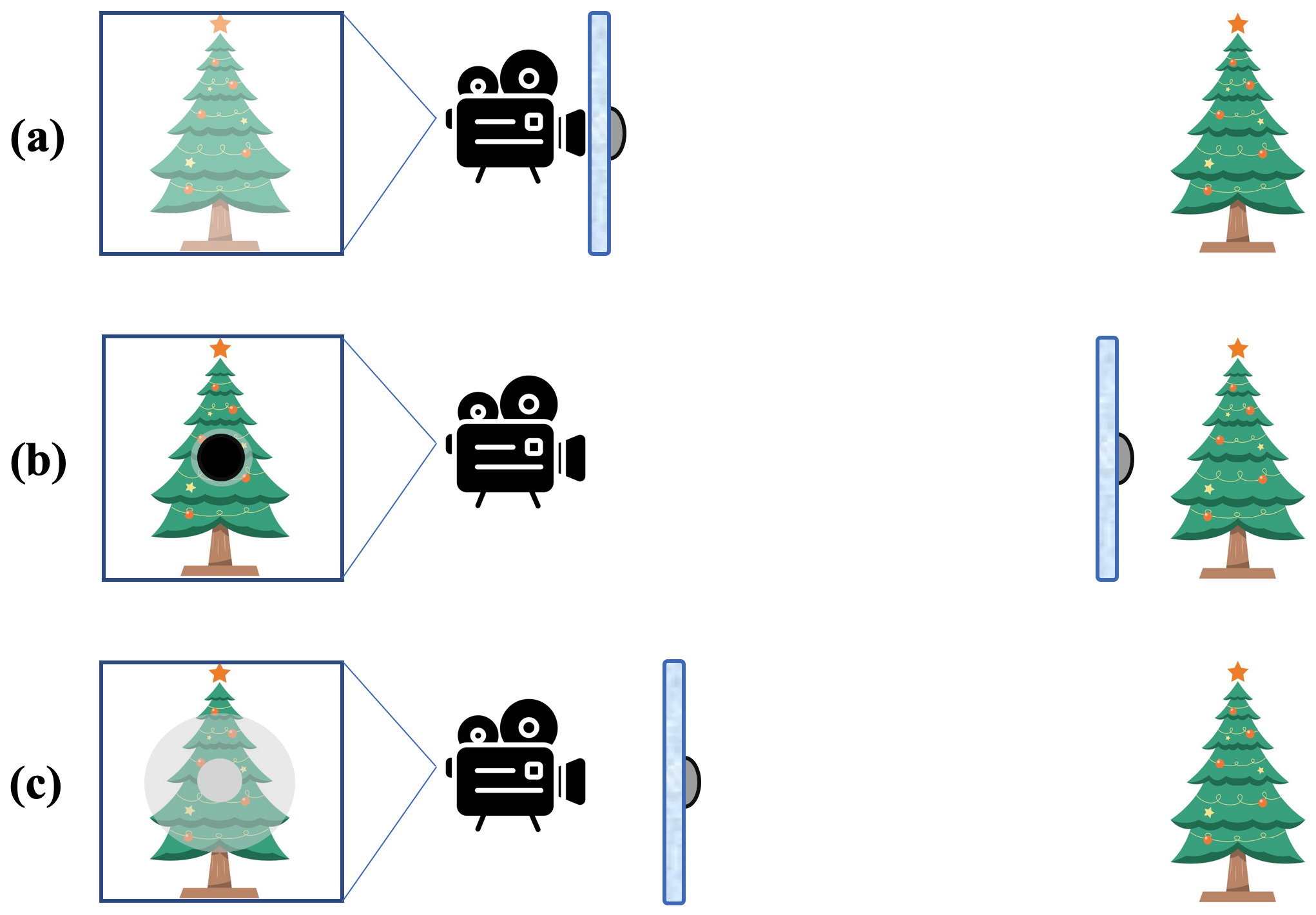}
\end{center}
    \caption{\textbf{Illustration of three shooting scenarios.}(a) Contaminants; (b) Obstructions; (c) Occlusions (Ours). }
    \label{fig: differences}
\end{figure}

\subsection{OROS Dataset Collection}

Training a deep model for occlusion removal requires abundant occlusion degraded images with ground truth clean images.  
Such dataset is absent in the literature, and generating synthetic data may introduce domain shift when the model is applied to real-world images~\cite{chi2021test}. To this end, we propose to collect occlusion/clean image pairs by following the observed occlusion formation model.

Taking the images pairs with and without glasses may cause misalignment due to refractive index mismatching \cite{li2022learning}. Thus, we utilize two pieces of identical clean camera glass (both have a light transmittance rate of 98.4\% and a thickness of 1 mm): one is adhered by occlusions; the other is kept clean. To avoid other misalignments, such as the camera's motion, atmospheric conditions and scene offsets, we collect the dataset using a tripod and remote control on constant ambient conditions and static backgrounds.

We use Sony A7RII with 85GM lens and Ricoh GR3X to take the images.
The distance between the glass and the camera lens ranges from 3-12 cm. 
We also choose environments where the glass filter will not produce reflection artifacts.

The proposed OROS dataset has two distinct degradations depending on their optical properties: thin occlusions (dirt) and thick occlusions (raindrops, muddy water and particles).
All images contain regions of partial occlusion, while images with thick occlusions also contain regions of complete occlusion. The thickness of the occlusions can be represented as 
$\mathbf{\alpha}$, the attenuation ratio of the occlusion layer \cite{baek2015muddy}. 
In our case, $\mathbf{\alpha \in  [0, 1]}$ and $\mathbf{\alpha_{o}} < \mathbf{\alpha_{m}} < \mathbf{\alpha_{p}}$, where $\mathbf{\alpha_{r}}$,  $\mathbf{\alpha_{m}}$, $\mathbf{\alpha_{p}}$ denote the $\mathbf{\alpha}$ of raindrops, muddy water and particles. 
In addition, we attached thick clods, stone fragments, flower/leaf fragments, and hand-drawn marks as particles.  
To ensure the preciseness of the dataset, we perform manual selection to ignore the defective image pairs. We elaborate on the dataset visually in the supplementary material.

\subsection{Theory of Occlusions}

In the subsection, we analyze the image formation model and the defocus imaging model based on our dataset. 
We present a unified degradation model to exhibit the double-degraded cases (the partial and  complete occlusion).

\noindent\textbf{Image Formation Model}: 
The image formation model is formed by placing an intermediate layer between the sensor and the target scene as in Fig.~\ref{fig: imaging}(a). 
The occlusion layer itself contributes some radiance to the camera, by either scattering light from other directions in the environment or reflecting light from its surface. 
We define $\mathbf{I_{s}(x, y)}$ and $\mathbf{I}$ as the radiance of the target scene and the image captured by the sensor; 
$\mathbf{\alpha(x, y) \in  [0, 1]}$ as the attenuation ratio of the occlusion layer (0 for completely blocked and 1 for completely passed); 
$\mathbf{I_{o}(x, y)}$ denotes the intensification part (the additional radiance from the occlusion layer itself). 
Thereby, the image formation model can be represented as: 
\begin{equation}\label{eq1}
\mathbf{I}= \mathbf{\alpha} \cdot \mathbf{I}_{s} * \mathbf{h} + (\mathbf{1 - \alpha}) \cdot \mathbf{I}_{o} * \mathbf{h},
\end{equation}
where $\mathbf{h(x,y,d)}$ is the point spread function (PSF) at the defocusposition $\mathbf{d}$ for the occlusion layer, and $*$ denotes image convolution.

The first term (attenuation) in Eq.~\ref{eq1} involves a convolution operator (pointwise multiplication), which allows the high-frequency components of $\mathbf{I}_{s}$ to be partially preserved in the captured image $\mathbf{I}$~\cite{gu2009removing}. The second term (intensification) in Eq.~\ref{eq1}  corresponds to the intensification, which comes from multiple sources: for thin occlusions, they are the scattering of the environmental lighting; for thick occlusions, they are the reflection of the occluders.
When the occlusion is very close to the lens, the defocus blur will be so large that both attenuation and intensification are always shown as low-frequency patterns in the degraded images.

 \renewcommand{\thefigure}{3}
\begin{figure}[!t]
  \centering
  \includegraphics[width=0.9\linewidth]{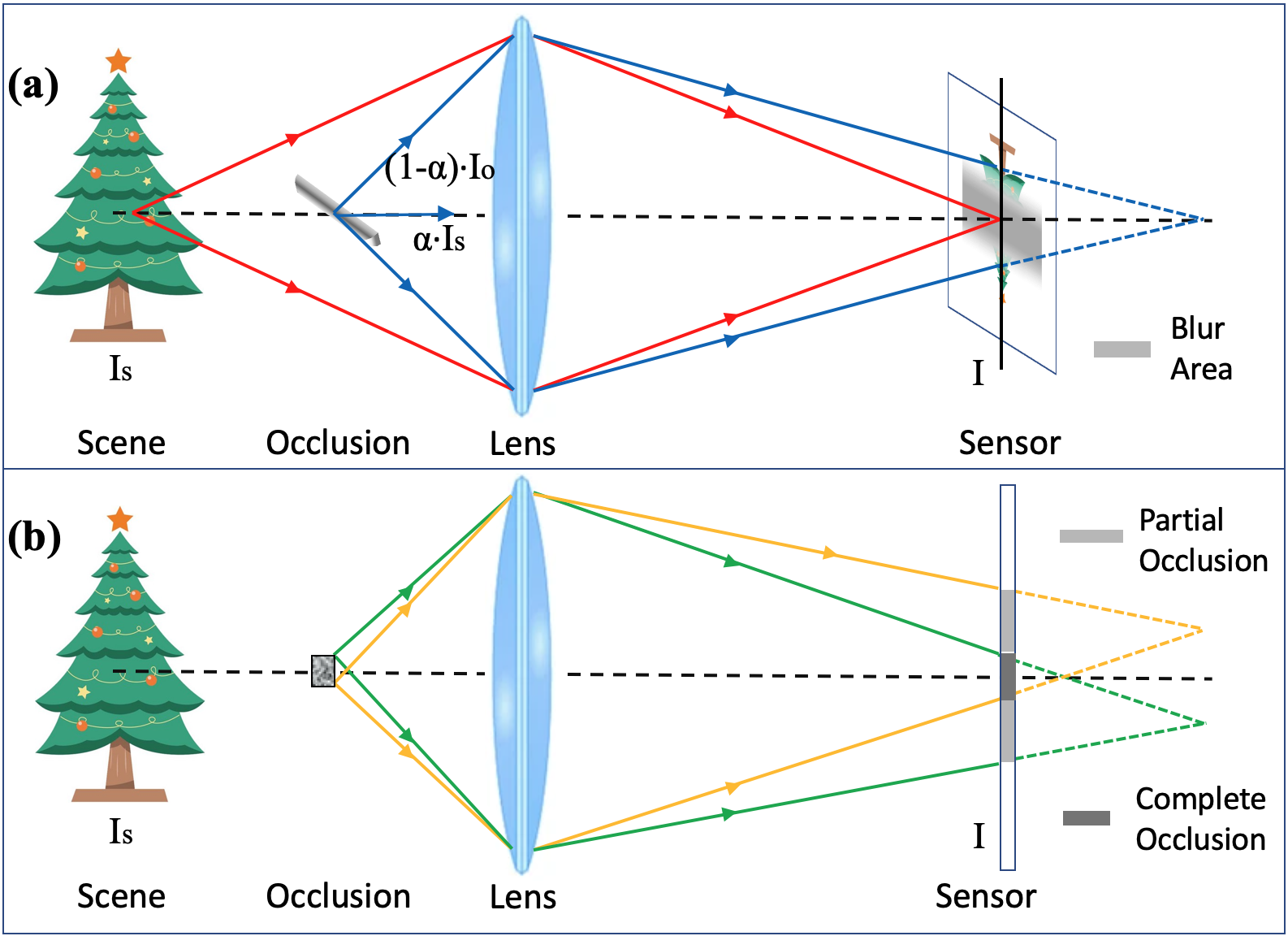}\\
 \vspace{0.2cm}
  \caption{\textbf{Models of occlusions.}(a) Image formation model; (b) Defocus imaging model.}
  \label{fig: imaging}

\end{figure}


\noindent\textbf{Defocus Imaging Model}: 
The sensor accumulates the radiance from the target scene, and defocuses the occlusion layer simultaneously
as in Fig.~\ref{fig: imaging}(b). 
Accordingly, the overall blur effect in the sensor is obtained by integrating the defocus blur produced by all regions of occlusions.
The light from the occlusion margin leads to defocus blur, which forms partial occlusions. 
The complete occlusion concentrates on the sensor's centre, while the partial one is around the complete occlusion.
In the thick occlusion margin, $\mathbf{\alpha(x, y)}$ gradually changes from 0 to 1 towards the center. The thick occlusion margin is a narrow band of a very small area similar to the occlusion region in the thin occlusions. We still define the margin as $\mathbf{I_{p}(x, y)}$, because of severe defocus blurring and over or under-exposure. The foreground of the thick occlusion appears as a region with uniform intensity. We define the such region as a constant $\mathbf{I}_{c}$$(x, y)$ and assume the depth of occlusion is also a constant.
Thereby, the following defocus imaging model in thick occlusions is:

\begin{equation} \label{eq4}
\mathbf{I} \approx  \mathbf{I}_{s} * \mathbf{h} + (\mathbf{\alpha - 1}) \cdot \mathbf{I}_{p} * \mathbf{h} + \mathbf{I}_{c}  * \mathbf{h},
\end{equation}
where $\mathbf{\alpha(x, y) \in  (0, 1)}$ and $\mathbf{h(x,y,d)}$ is the PSF at the defocus position $\mathbf{d}$. 
The formula contains a non-occlusion term, partial occlusion term and complete occlusion term.

\noindent\textbf{Unified Degraded Occlusion}: 
Combining the image formation model in Eq.~\ref{eq1} and the defocus imaging model in Eq.~\ref{eq4}, we derive the unified expression as:

\begin{equation} \label{eq5}
\mathbf{I} = \mathbf{I}_{N}  + (\mathbf{\alpha - 1}) \cdot \mathbf{I}_{P}  + \mathbf{\beta} \cdot \mathbf{I}_{C},
\end{equation}
where $\mathbf{I}_{N}$, $\mathbf{I}_{P}$ and $\mathbf{I}_{C}$ (non-occlusion, partial occlusion and complete occlusion) compose the image components in sensor; $\mathbf{\alpha(x, y) \in  [0, 1]}$; $\mathbf{\beta}$ denotes occlusions (0 for thin occlusions and 1 for thick occlusions).

Although the target scene radiance is partially preserved in the degraded images in Eq.~\ref{eq5}, the occlusion removal is still ill-posed in general because of both unknown $\mathbf{I}_{P}$ and $\mathbf{I}_{C}$.
Fortunately, we demonstrate that partial occlusion removal is the same to defocus deblurring, and complete occlusion removal is the same as inpainting in Eq.~\ref{eq5}.
Consequently, we could implement a multi-task network to remove occlusions in the double-degraded cases.

\section{The Proposed Method}

In this section, 
we present an all-in-one framework with self-supervised test-time adaptation for occlusion removal. The framework has two different tasks: the primary task takes input degraded images containing occlusions $\mathbf{I}_{P}$ and $\mathbf{I}_{C}$, and tries to predict a corresponding clean image $I^c$.
The self-supervised auxiliary task, particularly occlusion artifacts reconstruction, acts as a regularization in the joint training and prevents the network from overfitting.
On the other hand, the auxiliary task can learn optical occlusion features of each input $\mathbf{I}_{P}$ and $\mathbf{I}_{C}$ specifically, allowing model parameters to adapt to each input test image and assessing the primary task in producing a better result.

\subsection{Model Architecture}
In this subsection, we describe our two-branch network architecture:  the primary branch for the occlusion removal task and the auxiliary branch for the self-occlusion reconstruction task. Two tasks share the majority of the model's structure, which is an autoencoder.

\noindent\textbf{Primary Occlusion Removal Network}: 
The primary network input is an occlusion degraded image $I^d$, and the output is the clean counterpart $\hat{I}^c$, as shown in Fig.~\ref{fig:illustration}. 
We employ a multi-scale U-net structure~\cite{nah2017deep} with feature recurrence among different scales~\cite{park2020multi} and residual learning~\cite{he2016deep}. The feature maps in the decoder are passed to the encoder of the finer scale to contribute to the occlusion at finer scales. The detailed structure is provided in the supplementary.

\renewcommand{\thefigure}{4}
\begin{figure}
\begin{center}
    \includegraphics[width =0.9\linewidth]{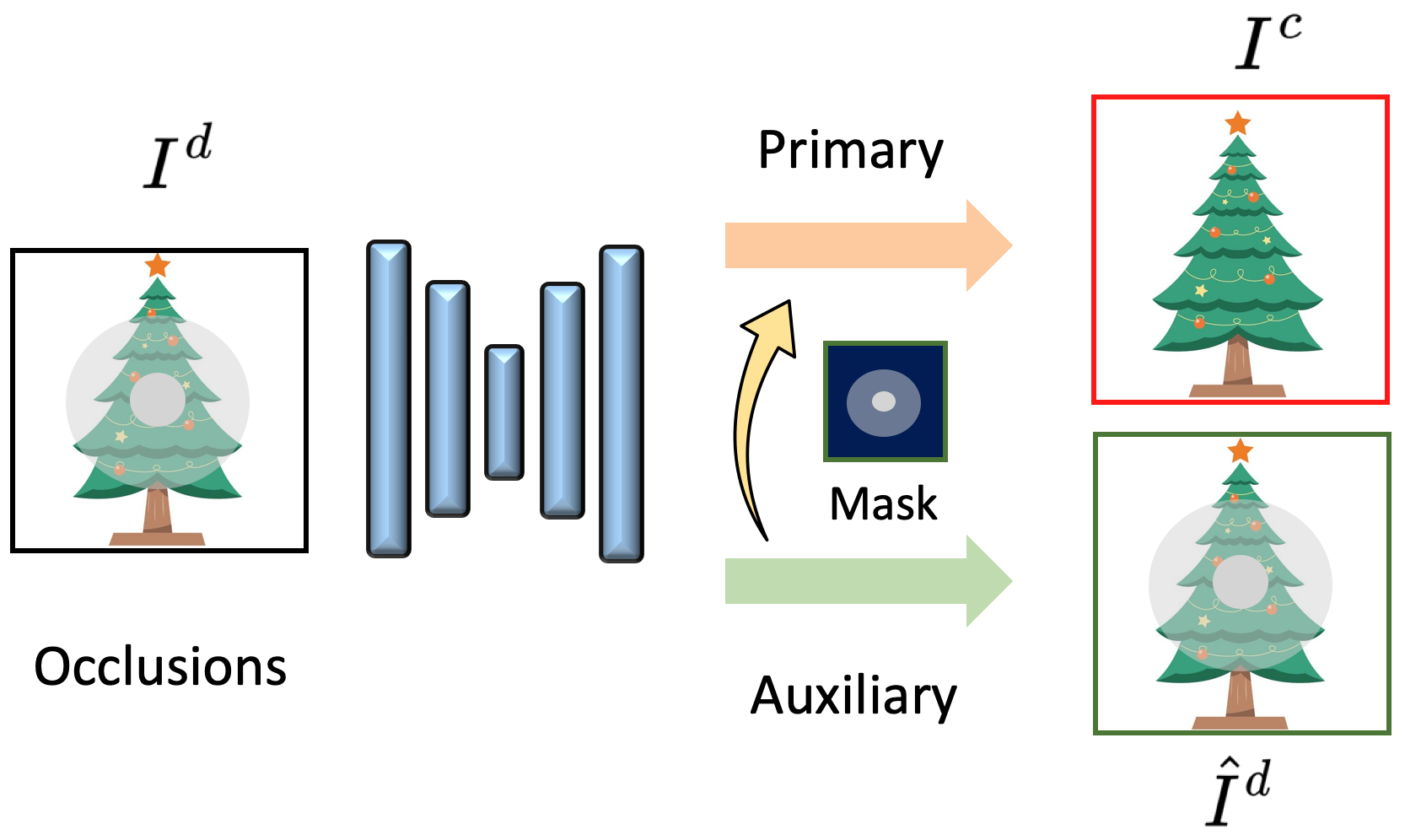}
\end{center}
     \vspace{-0.2cm}
    \caption{\textbf{Illustration of the proposed structure.} The input of the network is an occlusion-degraded image $I^d$; the primary branch tries to recover a clean image $I^c$; the auxiliary branch tries to reconstruct the occlusion-degraded image $\hat{I}^d$. We design an occlusion attention mask from auxiliary to primary. We illustrate details of network parameters in supplementary materials. }
    \label{fig:illustration}
\end{figure}

\renewcommand{\thefigure}{5}
\begin{figure}
\centering
\begin{subfigure}{0.2\linewidth}
	\includegraphics[width =\linewidth]{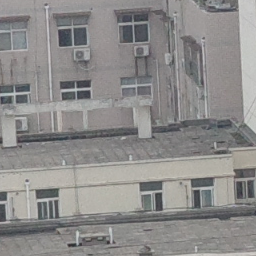}
	\subcaption*{GT}
\end{subfigure}
\begin{subfigure}{0.2\linewidth}
	\includegraphics[width =\linewidth]{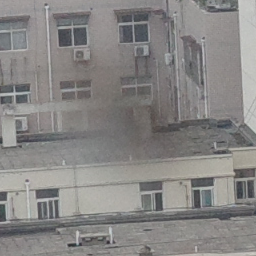}
	\subcaption*{Input}
\end{subfigure}
\begin{subfigure}{0.2\linewidth}
	\includegraphics[width =\linewidth]{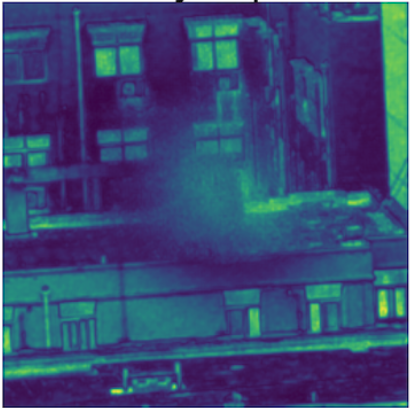}
	\subcaption*{Mask}
\end{subfigure}
    \caption{\textbf{Illustration of occlusion attention mask.} }
    \label{fig:mask}
\end{figure}

\noindent\textbf{Self-supervised Auxiliary Network}:
Generalization of unseen data is one of the typical issues associated with image restoration.  In our situation, there is an infinite number of occlusion kinds, making it impossible to incorporate them all into the training data. 
Thus, we employ auxiliary learning, which effectively improves the performance of the primary task on unseen data~\cite{liu2019self}. In our scenarios, we choose self-supervised reconstruction to enforce the auxiliary task to capture specific occlusion characteristics to elevate the primary occlusion removal task.  Adding such auxiliary task during training creates multi-task learning and serves as a regularization to improve the primary task. To further learn the occlusion-specific features ($\mathbf{I}_{P}$ and $\mathbf{I}_{C}$)
at test time, we incorporate test-time adaptation via self-reconstruction to update the model before inference. The auxiliary task accepts an occlusion-degraded image $I^d$, and reconstructs it ($\hat{I}^d$), as shown in Fig.~\ref{fig:illustration}.

\noindent\textbf{Occlusion Feature Mask}: 
 It is necessary to know what kind of information works in the occlusion removal process to identify the occluded regions (especially for various occluders), \eg the PSF kernel in \cite{shi2022seeing} and the relative layer motion in \cite{liu2020learning}. 
 In our network, we design an occlusion attention mask, mapping from the auxiliary branch to the primary branch.
 Such occlusion attention mask contributes to the learning process by attending the network to the occluded part, as illustrated by the feature map in Fig.~\ref{fig:mask}. More features are shown in supplementary materials.

\noindent\textbf{Differences with \cite{chi2021test}}: 
Compared to \cite{chi2021test}, our approach does not involve meta-learning at all. Such a learning paradigm aims to align the two conflicting learning objectives. However, in our case, the restoration is observed in local regions. Thus, reconstructing the clean part could be easy and enforces the network to identify and attend to the occluded part in Fig.~\ref{fig:mask}. Thus, our method does not need a heuristic design for occlusion identification. 
Our method also avoids cumbersome bi-level optimization and episodic-level task formulation. Waiving the inner loop also significantly reduced the training cost.

\subsection{Model Optimization}

Instead of having completely isolated branches\cite{sun2020test}, we pass the output of primary branch to the auxiliary branch. 
Consequently, our primary branch can still be adapted to any specific test images during the test time. Both tasks share most of the weights.

\noindent \textbf{Loss Functions}: We utilize the MS-SSIM + \emph{$l_{1}$} loss for two tasks as (multi-scale SSIM weighs SSIM  at different scales by the sensitivity of the HVS) \cite{zhao2016loss}:
\begin{gather}
\mathcal{L}_{P }=\alpha_{0} \cdot \mathcal{L}_{P}^{\text {MS-SSIM }}+(1-\alpha_{0}) \cdot G \cdot \mathcal{L}_{P }^{\ell_1},\\
\mathcal{L}_{A }=\alpha_{0} \cdot \mathcal{L}_{A}^{\text {MS-SSIM }}+(1-\alpha_{0}) \cdot G \cdot \mathcal{L}_{A}^{\ell_1},
\end{gather}
where $G$ denotes the Gaussian weights and the subscripts $P$ and $A$ denote the primary and auxiliary task. We set $\alpha_{0}=0.9$ empirically.

\noindent \textbf{Offline Training}: We define the parameters of the network as three terms: $\{ \theta, \theta_{p},  \theta_{a}\}$, where $\theta$ stands for the shared weights, $\theta_{p}$ denotes the task-specific weights for the primary occlusion removal branch, and $\theta_{a}$ denotes the reconstruction auxiliary branch. 
We define  $f_{p}^{\theta}$ and $f_{a}^{\theta}$ as functions of the primary occlusion removal branch and the reconstruction auxiliary branch, respectively. We obtain the predicted clean image of primary branch $\hat{I^{c}}$ and the reconstructed occlusion degraded image of auxiliary branch $\hat{I^{d}}$:
\begin{equation}
\small
\hat{I^{c}} = f_{p}^{\theta}(I^d; \theta_{p}), \hat{I_{d}} = f_{a}^{\theta}(\textit{$I^d$}; \theta_{a}, \theta_{p}).
\end{equation}
The auxiliary branch still contains $\theta_{p}$, since the auxiliary task obtains the output from the primary branch.
The process is necessary for the test-time adaptation \cite{chi2021test}.

We train our model jointly by combining the primary and auxiliary losses:

\begin{equation}
    \small
    \mathcal{L} = \alpha \cdot \mathcal{L}_{P}(I^c, \hat{I}^c;  \theta_{p}) + (1- \alpha) \cdot \mathcal{L}_{A}(I^d, \hat{I}^d;  \theta_{a}, \theta_{p}),
    \label{eq:joint_loss}
\end{equation}
where $\alpha=0.8$ in our case. 
In our paper, we define the model trained by Eq.~\ref{eq:joint_loss} called the base model. 
We could implement additional parameters update for specific input at test time.

\renewcommand{\thefigure}{6}
\begin{figure*}[t]
    \centering    
    \footnotesize
    \rotatebox{90}{\scriptsize{~~~~~~~~~~~~~~~~Dirt}}
    \begin{subfigure}{0.125\linewidth}
    \includegraphics[height=\linewidth, width =\linewidth]{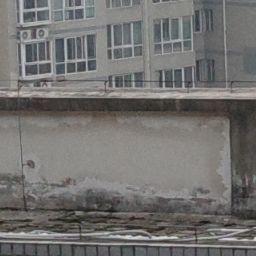}
    \end{subfigure}
    \begin{subfigure}{0.125\linewidth}
    \includegraphics[height=\linewidth, width =\linewidth]{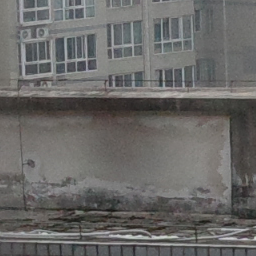}
    \end{subfigure}
    \begin{subfigure}{0.125\linewidth}
    \includegraphics[height=\linewidth, width =\linewidth]{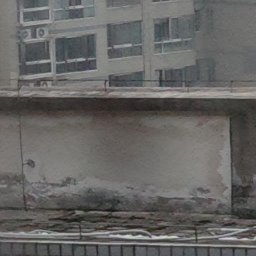}
    \end{subfigure}
    \begin{subfigure}{0.125\linewidth}
    \includegraphics[height=\linewidth, width =\linewidth]{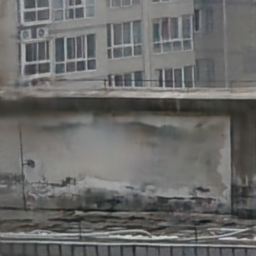}
    \end{subfigure}
    \begin{subfigure}{0.125\linewidth}
    \includegraphics[height=\linewidth, width =\linewidth]{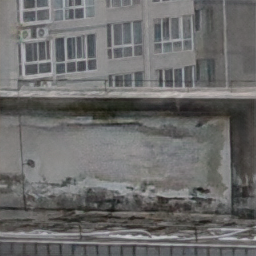}  
    \end{subfigure}
    \begin{subfigure}{0.125\linewidth}
    \includegraphics[height=\linewidth, width =\linewidth]{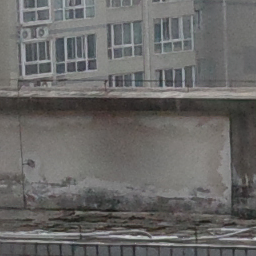}  
    \end{subfigure}
    \begin{subfigure}{0.125\linewidth}
    \includegraphics[height=\linewidth, width =\linewidth]{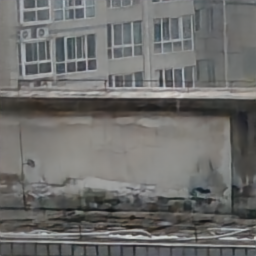}
    \end{subfigure}
    
    \rotatebox{90}{\scriptsize{~~~~~~~~~~~~Raindrops}}
    \begin{subfigure}{0.125\linewidth}
    \includegraphics[height=\linewidth, width =\linewidth]{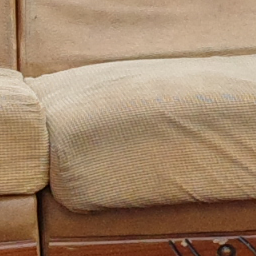}
    \end{subfigure}
    \begin{subfigure}{0.125\linewidth}
    \includegraphics[height=\linewidth, width =\linewidth]{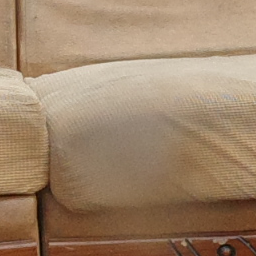}
    \end{subfigure}
    \begin{subfigure}{0.125\linewidth}
    \includegraphics[height=\linewidth, width =\linewidth]{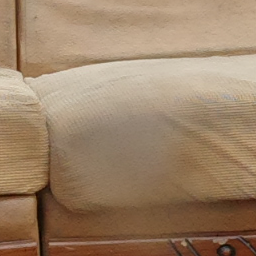}
    \end{subfigure}
    \begin{subfigure}{0.125\linewidth}
    \includegraphics[height=\linewidth, width =\linewidth]{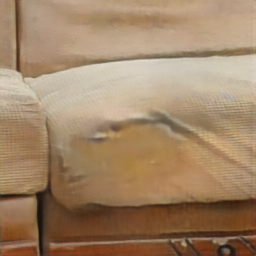}
    \end{subfigure}
    \begin{subfigure}{0.125\linewidth}
    \includegraphics[height=\linewidth, width =\linewidth]{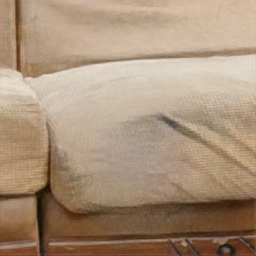}
    \end{subfigure}
    \begin{subfigure}{0.125\linewidth}
    \includegraphics[height=\linewidth, width =\linewidth]{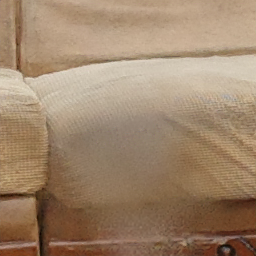}
    \end{subfigure}
    \begin{subfigure}{0.125\linewidth}
    \includegraphics[height=\linewidth, width =\linewidth]{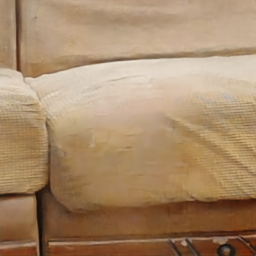}
    \end{subfigure}

    \rotatebox{90}{\scriptsize{~~~~~~~~Muddy Water}}
    \begin{subfigure}{0.125\linewidth}
    \includegraphics[height=\linewidth, width =\linewidth]{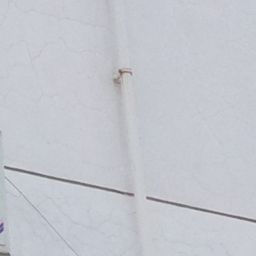}
    \end{subfigure}
    \begin{subfigure}{0.125\linewidth}
    \includegraphics[height=\linewidth, width =\linewidth]{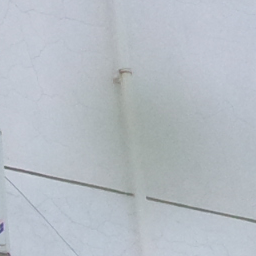}
    \end{subfigure}
    \begin{subfigure}{0.125\linewidth}
    \includegraphics[height=\linewidth, width =\linewidth]{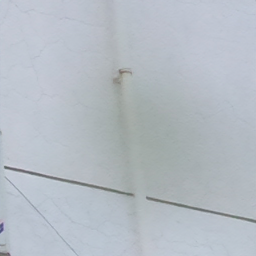}
    \end{subfigure}
    \begin{subfigure}{0.125\linewidth}
    \includegraphics[height=\linewidth, width =\linewidth]{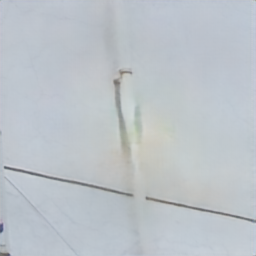}
    \end{subfigure}
    \begin{subfigure}{0.125\linewidth}
    \includegraphics[height=\linewidth, width =\linewidth]{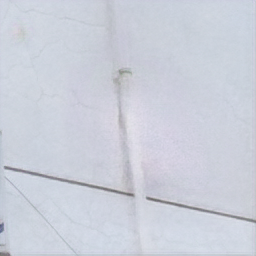}
    \end{subfigure}
    \begin{subfigure}{0.125\linewidth}
    \includegraphics[height=\linewidth, width =\linewidth]{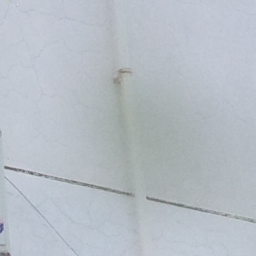}
    \end{subfigure}
    \begin{subfigure}{0.125\linewidth}
    \includegraphics[height=\linewidth, width =\linewidth]{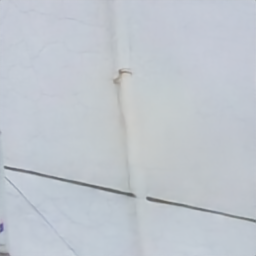}
    \end{subfigure}

    \rotatebox{90}{\scriptsize{~~~~~~~~~~~~~~~~~~~~Particles}}
    \begin{subfigure}{0.125\linewidth}
    \includegraphics[height=\linewidth, width =\linewidth]{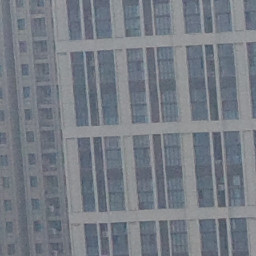}
    \caption{\footnotesize{GT}}
    \end{subfigure}
    \begin{subfigure}{0.125\linewidth}
    \includegraphics[height=\linewidth, width =\linewidth]{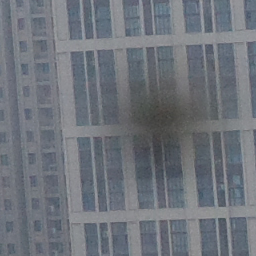}
    \caption{\footnotesize{Input}}
    \end{subfigure}
    \begin{subfigure}{0.125\linewidth}
    \includegraphics[height=\linewidth, width =\linewidth]{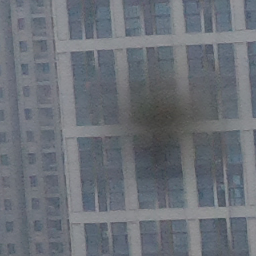}
    \caption{\footnotesize{PReNet}}
    \end{subfigure}
    \begin{subfigure}{0.125\linewidth}
    \includegraphics[height=\linewidth, width =\linewidth]{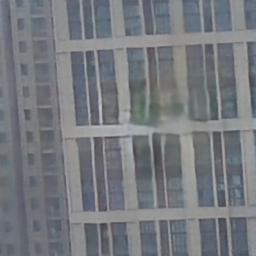}
    \caption{\footnotesize{U-net}}
    \end{subfigure}
    \begin{subfigure}{0.125\linewidth}
    \includegraphics[height=\linewidth, width =\linewidth]{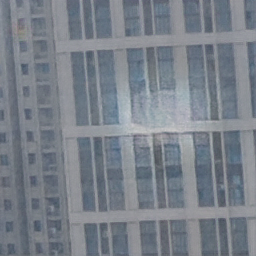}
    \caption{\footnotesize{DeblurGAN-v2}}  
    \end{subfigure}
    \begin{subfigure}{0.125\linewidth}
    \includegraphics[height=\linewidth, width =\linewidth]{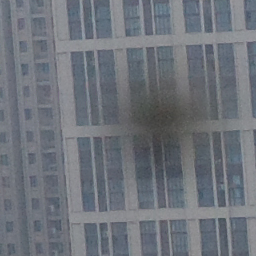}
    \caption{\footnotesize{CTSDG}}  
    \end{subfigure}
    \begin{subfigure}{0.125\linewidth}
    \includegraphics[height=\linewidth, width =\linewidth]{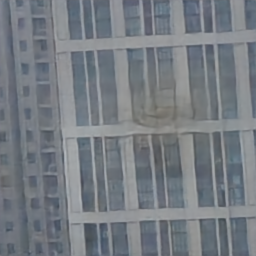}
    \caption{\footnotesize{Ours}}
    \end{subfigure}
    \caption{\textbf{Qualitative comparison with state-of-the-art approaches on our OROS dataset.} Our method yields sharper results than the state-of-the-art approaches.}
    \label{fig: compare sota}
\end{figure*}

\begin{table}[t!]
\small
\centering
\setlength{\tabcolsep}{2.2pt} 
\begin{tabular}{lcccc}
\toprule
\multirow{2}{*}{\diagbox [width=8em,trim=l] {Methods}{Dataset}} & 
\multicolumn{2}{c}{OROS}  \\
\cmidrule(r){2-3} 
& PSNR & SSIM \\
\midrule
Degraded  & 23.31  & 0.846   \\
PReNet \cite{ren2019progressive}  & 22.80  & 0.835   \\
U-net \cite{wu2020dense}  & 24.70  &  0.846  \\ 
DeblurGAN-v2 \cite{kupyn2019deblurgan} & 28.18  & 0.857  \\ 
CTSDG \cite{guo2021image} & 24.96  & 0.794 \\
\midrule
Ours& \textbf{29.91}  & \textbf{0.879}  \\
\bottomrule
\end{tabular}
\caption{\textbf{Comparison of our model with state-of-the-art on our OROS dataset. } All the networks are re-trained on OROS dataset. Our model outperforms the existing SOTA.}

\label{tab: compare SOTA}
\end{table}

\noindent\textbf{Test-time Adaption}:  The proposed method adopts a self-supervised auxiliary task to further update the trained model towards each test image by learning its unique occlusion properties.
We learn the model parameters $\theta$, 
and the adapted parameter $\tilde{\theta}$ is obtained by auxiliary loss update in the testing phase. 
Then $\tilde{\theta}$ is adopted to remove occlusions in the degraded image.

\section{Experiments}

In this section, we conduct extensive experiments on OROS to demonstrate the effectiveness of our scheme

\subsection{Implementation Details}

\noindent\textbf{Data Preparation and Metrics}: We train our network on the OROS training dataset, which consists of 2637 training pairs. Our testing dataset consists of 333 images: thin occlusions (dirt) and thick occlusions (raindrops, muddy water and occluders).
For the whole dataset, the image resolution is $256 \times 256$. We adopt PSNR and SSIM~\cite{wang2004image} as the evaluation metrics.

\noindent\textbf{Implementation and Training Details}: Pytorch is utilized for all implementations. We trained the network for 1300 iterations using Adam~\cite{kingma2014adam} as the optimizer with a $\beta_1$ of 0.5 and a $\beta_2$ of  0.999. The learning rate is set to 0.0001 and batch size is set to 30. We perform 6 gradient updates for each test image during the test-time adaptation. The learning rate for gradient update is fixed to 0.000006. We scaled all pixel values of both training and testing images to [-1, 1],  and LeakyReLU~\cite{xu2015empirical} with a slope of 0.1 is used as the activation function. We perform model training, testing and adaption on a Nvidia 3090 GPU.

\begin{table}[t!]
\footnotesize
\centering
\setlength{\tabcolsep}{1.1pt} 
\begin{tabular}{lcccccccc}
\toprule
\multirow{2}{*}{\diagbox [width=6em,trim=l] {Methods}{Dataset}} & 
\multicolumn{2}{c}{Dirt} & 
\multicolumn{2}{c}{Raindrops} &
\multicolumn{2}{c}{Muddy Water} & 
\multicolumn{2}{c}{Particles} \\
\cmidrule(r){2-3} \cmidrule(r){4-5} \cmidrule(r){6-7} \cmidrule(r){8-9} 
& PSNR & SSIM & PSNR & SSIM & PSNR & SSIM & PSNR & SSIM\\
\midrule
Degraded  & 25.79 & 0.861 & 24.83 & 0.827 & 24.25 & 0.833 & 23.07 & 0.788\\
PReNet\cite{ren2019progressive}  & 25.10 & 0.850 & 24.52 & 0.817 & 23.51 & 0.809 & 19.62 & 0.836\\
U-net\cite{wu2020dense}  & 24.19 & 0.834 & 24.73 & 0.826 & 25.48 & 0.827 & 24.92 & 0.872\\
DeblurGAN-v2\cite{kupyn2019deblurgan}  & 28.52 & 0.856 & 28.28 & 0.832 & 27.12 & 0.835 & 28.15 & 0.877\\
CTSDG \cite{guo2021image} & 26.38 & 0.805 & 26.22 & \textbf{0.872} & 24.94 & 0.758 & 23.07 & 0.788 \\
\midrule
Ours   & \textbf{29.93} & \textbf{0.878} & \textbf{29.86} & 0.852 & \textbf{29.53} & \textbf{0.852} & \textbf{30.18} & \textbf{0.899}\\   
\bottomrule
\end{tabular}
\caption{\textbf{Comparison on four sub-datasets of our OROS dataset(dirt, raindrops, muddy water and particles)}. Our two branches model obtained by optimizing Eq.~\ref{eq:joint_loss} outperforms current state-of-the-art.
  }
\label{tab:four sub-datasets}
\end{table}

\subsection{Comparison with the State-of-the-arts}

We provide objective
and subjective quality comparison on OROS dataset with state-of-the-art learning-based methods: PReNet (deraining) \cite{ren2019progressive}, U-net (deblurring) \cite{wu2020dense}, DeblurGAN-v2 (deblurring) \cite{kupyn2019deblurgan}, 
CTSDG (inpainting) \cite{guo2021image}. 
For each of these methods,  we retrain the released official model on our OROS dataset. The comparison is faithful and fair by finding their optimal hyperparameters.

\noindent\textbf{Quantitative Evaluation}: The objective performance evaluation of PSNR and SSIM on OROS dataset is shown in Tab. \ref{tab: compare SOTA}.
It can be found that, our method achieves the best PSNR and SSIM performance on OROS dataset.
We also report the results on four sub-datasets of OROS dataset in Tab. \ref{tab:four sub-datasets}. 
It can be seen that ours outperforms the state-of-the-art methods except for CTSDG of Raindrops on the OROS dataset.
In addition, DeblurGAN-v2 produces a plausible hypothesis for the missing region conditioned on its surroundings\cite{pathak2016context}.    
Although DeblurGAN-v2 boosts deblurring and inpainting performance simultaneously, it still fails quickly in cases where patterns are non-repetitive or unique.
In contrast, our method learns optical occlusion properties of partial and complete occlusion by the auxiliary task with time-adaption.
Our method, adapted to various occlusion types, achieves the best performance on most sub-datasets.
These comparison results demonstrate the superior performance of our proposed network.

\noindent\textbf{Qualitative Evaluation}: 
The visual comparison results on four sub-datasets of OROS are illustrated in Fig. \ref{fig: compare sota}.
As can be seen, our method is considerably more effective in removing occlusions compared to the state-of-the-art methods.
PReNet \cite{ren2019progressive} performs poorly as a deraining network, even on the Raindrops sub-dataset.
U-net \cite{wu2020dense} performs well on the Muddy Water sub-dataset as a deblurring network due to the similar training distribution. However, U-net can not recover the underlying information from the occlusion images with multiple degradations.
CTSDG \cite{guo2021image} only performs well on the Raindrops sub-dataset.
Next, we compare the best two methods (ours and DeblurGAN-v2) in detail:
(1) On the Dirt sub-dataset, the predicted clean image of DeblurGAN-v2 contains much black texture, while ours yields a cleaner result;
(2) On the Raindrop sub-dataset, the output of DeblurGAN-v2 shows many black stripes. In contrast, we exhibit an object with smoothness;
(3) On the Muddy Water sub-dataset, although the result of DeblurGAN-v2 displays the object with more explicit edge boundaries and structures, it still has a similar overexposed effect. 
Instead, ours has a balance between luminance and details;
(4) On the Occluders sub-dataset, DeblurGAN-v2 still produces overexposed artifacts.
In contrast, our method produces a visually pleasing result.
In comparison, as the occlusion distribution shift, the compared methods suffer from poor generalization with incomplete occlusion removal and artifacts. 
Instead, our method can remove occlusions with fewer artifacts and obtain the best visual quality, because our model learns to adapt effectively to each test image's specific occlusion information.
Our method considerably boosts state-of-the-art occlusion removal performance.

\begin{table}[t!]
\footnotesize
\centering
\setlength{\tabcolsep}{1.1pt} 
\begin{tabular}{lcccccccc}
\toprule
\multirow{2}{*}{\diagbox [width=6em,trim=l] {Methods}{Dirt}} & 
\multicolumn{2}{c}{Unseen Ones} \\
\cmidrule(r){2-3} 
& PSNR & SSIM \\
\midrule
DeblurGAN-v2\cite{kupyn2019deblurgan} & 25.57 & 0.802 \\
CTSDG \cite{guo2021image} &  25.29 & 0.814 \\
\midrule

Ours (no update)  &  29.39 & 0.871 \\
Ours (3 updates) &  \textbf{29.67} & \textbf{0.876} \\   
\bottomrule
\end{tabular}
\caption{\textbf{Comparison on unseen ones on Dirt.}
  }
\label{tab:unseen}
\end{table}

\noindent\textbf{Results on Unseen
Ones}: The test-time adaptation mechanism can help the network deal with unseen  occlusions. To better verify this point, we experiment by training the network with specific sub-datasets of the OSOR but testing it on other sub-datasets. 
For example, we train the model using
Raindrops, Muddy Water, and Particles sub-datasets and
test it on the Dirt sub-dataset in Tab. \ref{tab:unseen}. This setting imitates real-world cases with the occlusion kind never seen in model training, and thus can validate the benefits of the test-time adaptation mechanism.
We implement 20 updates and achieve the best quantitative performance in 3 updates. 
It validates the benefits of the test-time adaptation mechanism for better
generalization.

\noindent\textbf{Computational Cost}:
We test the speed of running a $256\times256$ image on a RTX3090 GPU. 
Our method requires 0.17s w/o test-time adaptation and 1.30s w/ 6 gradient updates. However, since DeblurGAN-v2 requires additional features from a pre-trained network, it needs 1.68s. Through comparison, our method is simple yet efficient even with adaptive steps.

\renewcommand{\thefigure}{7}

\begin{figure}
    \centering
    \rotatebox{90}{\scriptsize{~~~~~Dirt}}
	\begin{subfigure}{0.128\linewidth}
		\includegraphics[width =\linewidth]{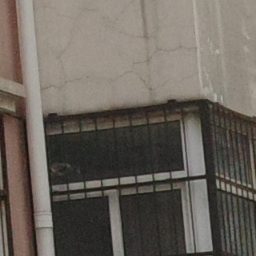}
	\end{subfigure}
	\begin{subfigure}{0.128\linewidth}
		\includegraphics[width =\linewidth]{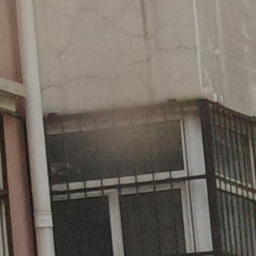}
	\end{subfigure}
	\begin{subfigure}{0.128\linewidth}
		\includegraphics[width =\linewidth]{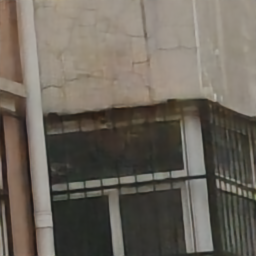}
	\end{subfigure}
	\begin{subfigure}{0.128\linewidth}
		\includegraphics[width =\linewidth]{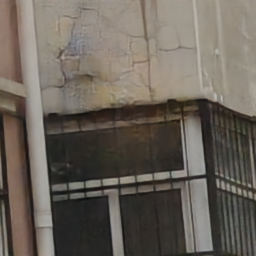}
	\end{subfigure}
	\begin{subfigure}{0.128\linewidth}
		\includegraphics[width =\linewidth]{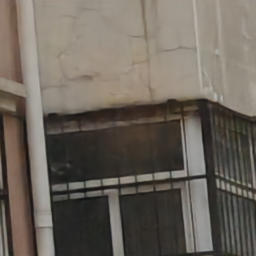}
	\end{subfigure}
 	\begin{subfigure}{0.128\linewidth}
		\includegraphics[width =\linewidth]{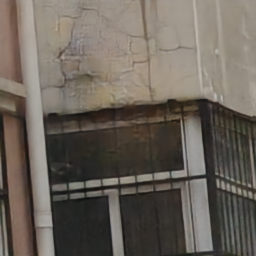}
	\end{subfigure}
    \begin{subfigure}{0.128\linewidth}
		\includegraphics[width =\linewidth]{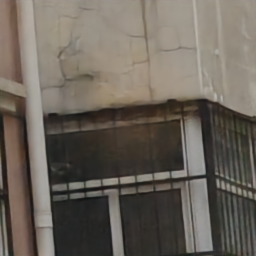}
	\end{subfigure}

    \rotatebox{90}{\scriptsize{Raindrops}}
	\begin{subfigure}{0.128\linewidth}
		\includegraphics[width =\linewidth]{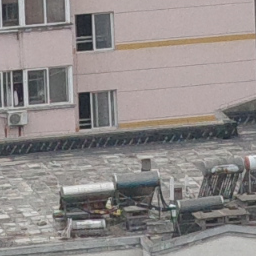}
	\end{subfigure}
	\begin{subfigure}{0.128\linewidth}
		\includegraphics[width =\linewidth]{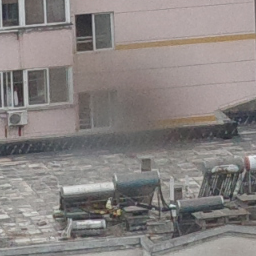}
	\end{subfigure}
	\begin{subfigure}{0.128\linewidth}
		\includegraphics[width =\linewidth]{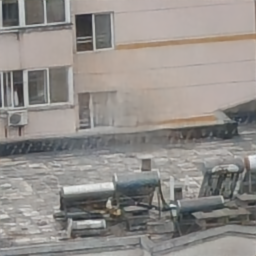}
	\end{subfigure}
	\begin{subfigure}{0.128\linewidth}
		\includegraphics[width =\linewidth]{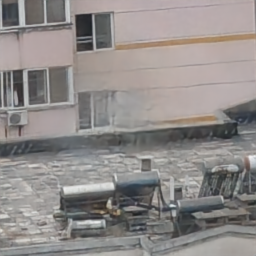}
	\end{subfigure}
	\begin{subfigure}{0.128\linewidth}
		\includegraphics[width =\linewidth]{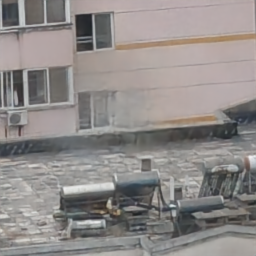}
	\end{subfigure}
    \begin{subfigure}{0.128\linewidth}
		\includegraphics[width =\linewidth]{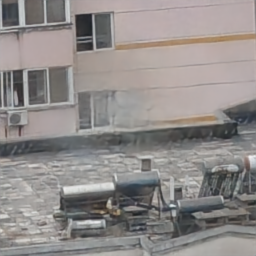}
	\end{subfigure}
	\begin{subfigure}{0.128\linewidth}
		\includegraphics[width =\linewidth]{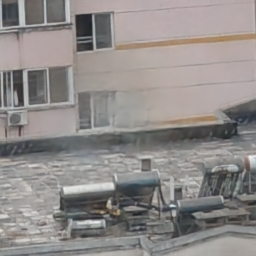}
	\end{subfigure}

\rotatebox{90}{\scriptsize{~~Muddy}}
\begin{subfigure}{0.128\linewidth}
	\includegraphics[width =\linewidth]{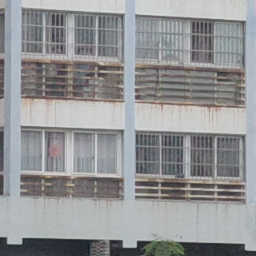}
\end{subfigure}
\begin{subfigure}{0.128\linewidth}
	\includegraphics[width =\linewidth]{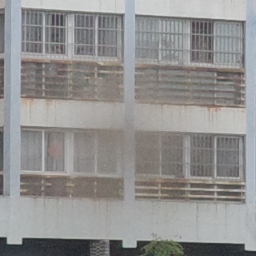}
\end{subfigure}
\begin{subfigure}{0.128\linewidth}
	\includegraphics[width =\linewidth]{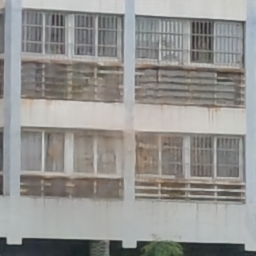}
\end{subfigure}
\begin{subfigure}{0.128\linewidth}
	\includegraphics[width =\linewidth]{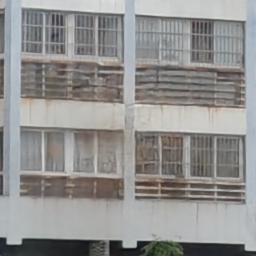}
\end{subfigure}
\begin{subfigure}{0.128\linewidth}
	\includegraphics[width =\linewidth]{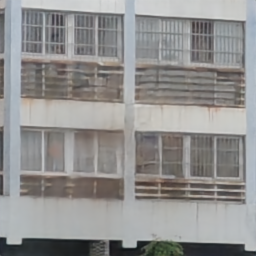}
\end{subfigure}
\begin{subfigure}{0.128\linewidth}
	\includegraphics[width =\linewidth]{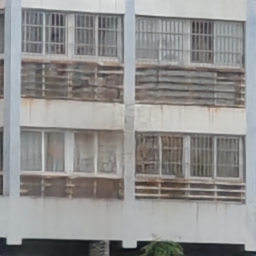}
\end{subfigure}
\begin{subfigure}{0.128\linewidth}
	\includegraphics[width =\linewidth]{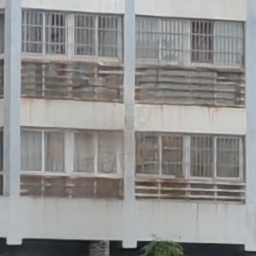}
\end{subfigure}

\rotatebox{90}{\scriptsize{~~~~~~~~~~~~Particles}}
\begin{subfigure}{0.128\linewidth}
	\includegraphics[width =\linewidth]{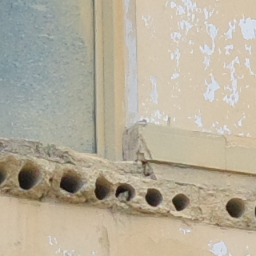}
	\subcaption*{GT}
\end{subfigure}
\begin{subfigure}{0.128\linewidth}
	\includegraphics[width =\linewidth]{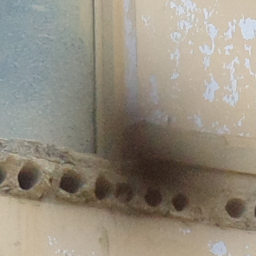}
	\subcaption*{Input}
\end{subfigure}
\begin{subfigure}{0.128\linewidth}
	\includegraphics[width =\linewidth]{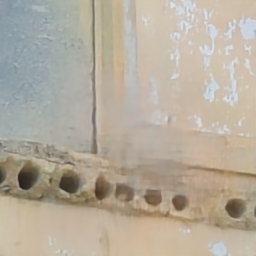}
	\subcaption*{(a)}
\end{subfigure}
\begin{subfigure}{0.128\linewidth}
	\includegraphics[width =\linewidth]{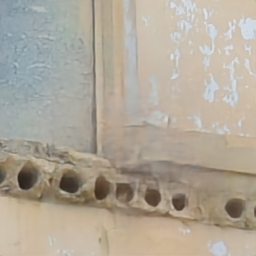}
	\subcaption*{(b)}
\end{subfigure}
\begin{subfigure}{0.128\linewidth}
	\includegraphics[width =\linewidth]{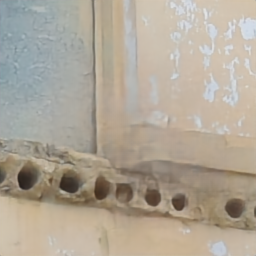}
	\subcaption*{(c)}
\end{subfigure}
\begin{subfigure}{0.128\linewidth}
	\includegraphics[width =\linewidth]{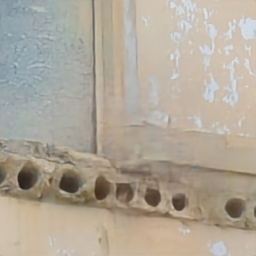}
	\subcaption*{(d)}
\end{subfigure}
\begin{subfigure}{0.128\linewidth}
	\includegraphics[width =\linewidth]{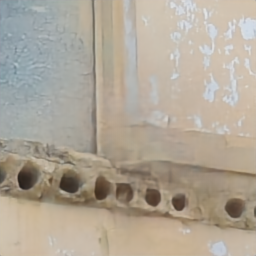}
	\subcaption*{(e)}
\end{subfigure}
    \caption{\textbf{Qualitative comparison with different network structures.} (a) Primary; (b) Primary + Auxiliary; (c) Primary + Auxiliary + Update; (d) Primary + Auxiliary + Mask; (e) Primary + Auxiliary + Update + Mask.}
    \label{fig:ablation}
\end{figure}

\subsection{Ablation Studies}
We conduct ablation experiments to further investigate the effect of various network structures and the number of gradient updates on the proposed OROS dataset.

\noindent\textbf{Network Structures}: We evaluate the effect of each component of our network structures.
We train five components of our network: 
(a) primary branch without the auxiliary branch; 
(b) our base model with the primary branch and the auxiliary branch; 
(c) our base model with $\mathbf{n}$ times test-time adaption parameters update. In our best case on OROS dataset, we set $\mathbf{n}$ as 4;
(d) our base model with the occlusion attention mask;
(e) our base model with test-time adaption update and occlusion attention mask.
Tab.~\ref{tab:ablation studies for conment} shows the quantitative comparison of different network structures of our model on the OROS dataset.
Compared with (a), (b) has the better PSNR and SSIM. 
The reason is that multiple tasks (primary branch + auxiliary branch) are more challenging to train than a single task.
Compared with others, the case (e) gives a performance boost greatly in PSNR and SSIM.
The qualitative evaluation in Fig.~\ref{fig:ablation} also shows the effectiveness of auxiliary task, test-time adaption and occlusion attention mask.
In some cases, the combination of primary and auxiliary branches shows more artifacts than the primary branch. 
Instead, the case (e) significantly boosts the occlusion removal performance by updating the parameters for each test image.

\begin{table}[t!]
\footnotesize
\centering
\setlength{\tabcolsep}{4pt} 
\begin{tabular}{lcc}
\toprule
\multirow{1}{*}{Network Structures} & 
\multicolumn{1}{c}{PSNR} & 
\multicolumn{1}{c}{SSIM} \\
\midrule
Primary & 29.16 & 0.869  \\
Primary + Auxiliary & 29.42 & 0.872  \\
Primary + Auxiliary + Update & 29.82  & 0.875   \\
Primary + Auxiliary + Mask & 29.50 & 0.873  \\
Primary + Auxiliary + Update + Mask &  \textbf{29.91} & \textbf{0.879}   \\
\bottomrule
\end{tabular}
\caption{\textbf{Ablation studies on network structures.} Including the auxiliary-learning through self occlusion reconstruction enhance the performance of primary task. As gradient updates and the attention mask are utilized, the performance of the primary task improves more.}
 
\label{tab:ablation studies for conment}
\end{table}

\renewcommand{\thefigure}{8}
\begin{figure}[t]
    \centering    
    \small
    \rotatebox{90}{\scriptsize{~~~~~Dirt}}
    \begin{subfigure}{0.128\linewidth}
    \includegraphics[height=\linewidth, width =\linewidth]{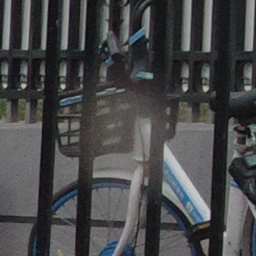}
    \end{subfigure}
    \begin{subfigure}{0.128\linewidth}
    \includegraphics[height=\linewidth, width =\linewidth]{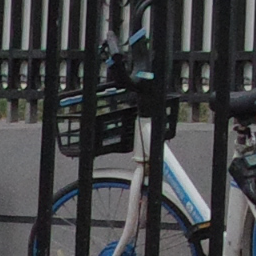}
    \end{subfigure}
    \begin{subfigure}{0.128\linewidth}
    \includegraphics[height=\linewidth, width =\linewidth]{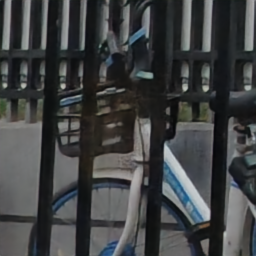}
    \end{subfigure}
    \begin{subfigure}{0.128\linewidth}
    \includegraphics[height=\linewidth, width =\linewidth]{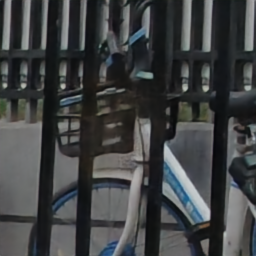}
    \end{subfigure}
    \begin{subfigure}{0.128\linewidth}
    \includegraphics[height=\linewidth, width =\linewidth]{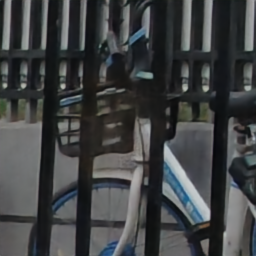}
    \end{subfigure}
    \begin{subfigure}{0.128\linewidth}
    \includegraphics[height=\linewidth, width =\linewidth]{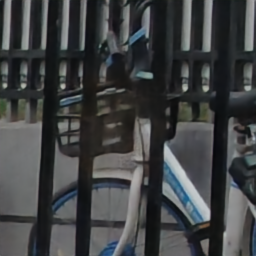}
    \end{subfigure}
    \begin{subfigure}{0.128\linewidth}
    \includegraphics[height=\linewidth, width =\linewidth]{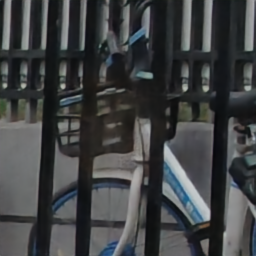}
    \end{subfigure}

     \rotatebox{90}{\scriptsize{Raindrops}}
    \begin{subfigure}{0.128\linewidth}
    \includegraphics[height=\linewidth, width =\linewidth]{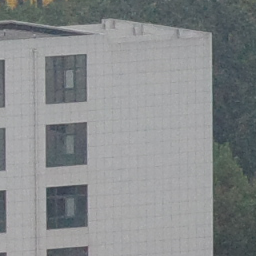}
    \end{subfigure}
    \begin{subfigure}{0.128\linewidth}
    \includegraphics[height=\linewidth, width =\linewidth]{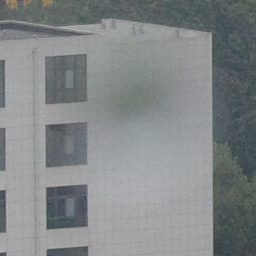}
    \end{subfigure}
    \begin{subfigure}{0.128\linewidth}
    \includegraphics[height=\linewidth, width =\linewidth]{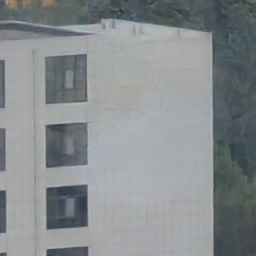}
    \end{subfigure}
    \begin{subfigure}{0.128\linewidth}
    \includegraphics[height=\linewidth, width =\linewidth]{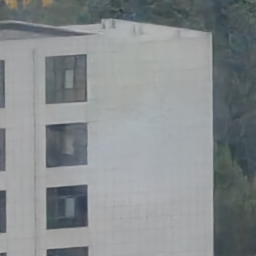}
    \end{subfigure}
    \begin{subfigure}{0.128\linewidth}
    \includegraphics[height=\linewidth, width =\linewidth]{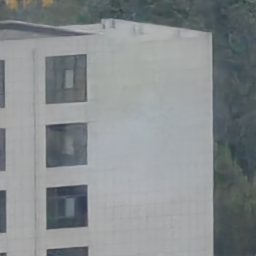}
    \end{subfigure}
    \begin{subfigure}{0.128\linewidth}
    \includegraphics[height=\linewidth, width =\linewidth]{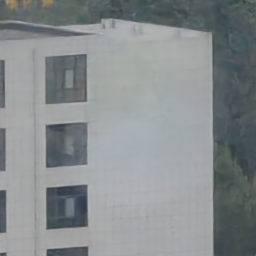}
    \end{subfigure}
    \begin{subfigure}{0.128\linewidth}
    \includegraphics[height=\linewidth, width =\linewidth]{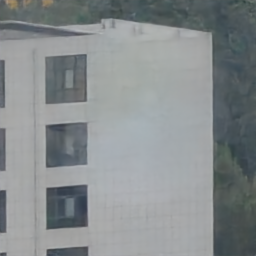}
    \end{subfigure}
    
    \rotatebox{90}{\scriptsize{~~~Muddy }}
    \begin{subfigure}{0.128\linewidth}
    \includegraphics[height=\linewidth, width =\linewidth]{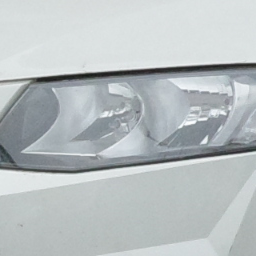}
    \end{subfigure}
    \begin{subfigure}{0.128\linewidth}
    \includegraphics[height=\linewidth, width =\linewidth]{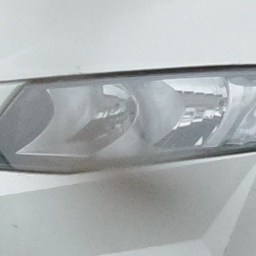}
    \end{subfigure}
    \begin{subfigure}{0.128\linewidth}
    \includegraphics[height=\linewidth, width =\linewidth]{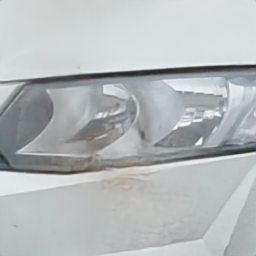}
    \end{subfigure}
    \begin{subfigure}{0.128\linewidth}
    \includegraphics[height=\linewidth, width =\linewidth]{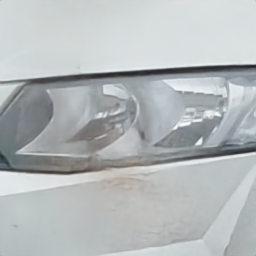}
    \end{subfigure}
    \begin{subfigure}{0.128\linewidth}
    \includegraphics[height=\linewidth, width =\linewidth]{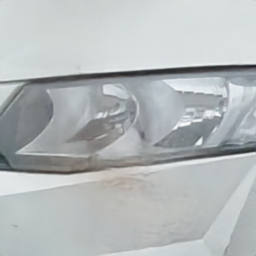}
    \end{subfigure}
    \begin{subfigure}{0.128\linewidth}
    \includegraphics[height=\linewidth, width =\linewidth]{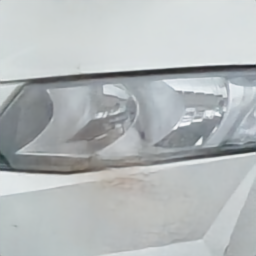}
    \end{subfigure}
    \begin{subfigure}{0.128\linewidth}
    \includegraphics[height=\linewidth, width =\linewidth]{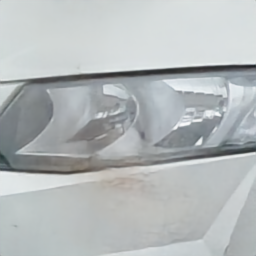}
    \end{subfigure}
    
    \rotatebox{90}{\scriptsize{~~~~~~~~~~~~Particles}}
    \begin{subfigure}{0.128\linewidth}
    \includegraphics[height=\linewidth, width =\linewidth]{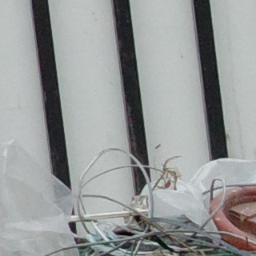}
    \subcaption*{GT}
    \end{subfigure}
    \begin{subfigure}{0.128\linewidth}
    \includegraphics[height=\linewidth, width =\linewidth]{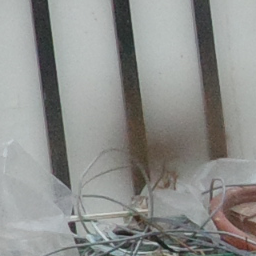}
    \subcaption*{Input}
    \end{subfigure}
    \begin{subfigure}{0.128\linewidth}
     \includegraphics[height=\linewidth, width =\linewidth]{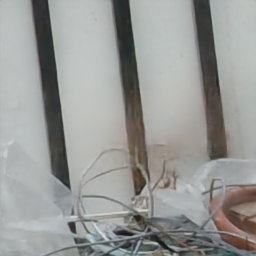}
    \subcaption*{0}
    \end{subfigure}
    \begin{subfigure}{0.128\linewidth}
    \includegraphics[height=\linewidth, width =\linewidth]{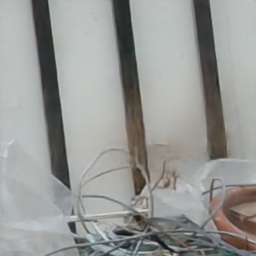}
    \subcaption*{2}
    \end{subfigure}
    \begin{subfigure}{0.128\linewidth}
     \includegraphics[height=\linewidth, width =\linewidth]{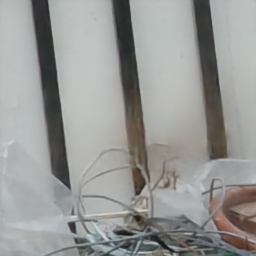}
    \subcaption*{4}
    \end{subfigure}
    \begin{subfigure}{0.128\linewidth}
    \includegraphics[height=\linewidth, width =\linewidth]{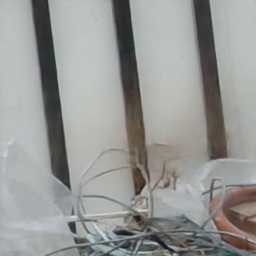}
    \subcaption*{6}
    \end{subfigure}
    \begin{subfigure}{0.128\linewidth}
    \includegraphics[height=\linewidth, width =\linewidth]{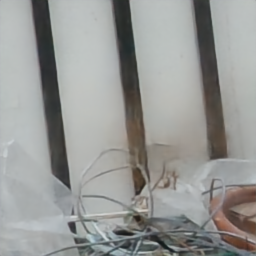}
    \subcaption*{8}
    \end{subfigure}

    \caption{\textbf{Visual illustration of different test time updates on our OROS dataset.} With the test-time adaptation, the artifacts and incomplete blur removal suffered from distribution shift are resolved.} 
    \label{fig: n-update}
\end{figure}

\renewcommand{\thefigure}{9}
\begin{figure}[t]
\centering
    \includegraphics[width =0.45\linewidth]{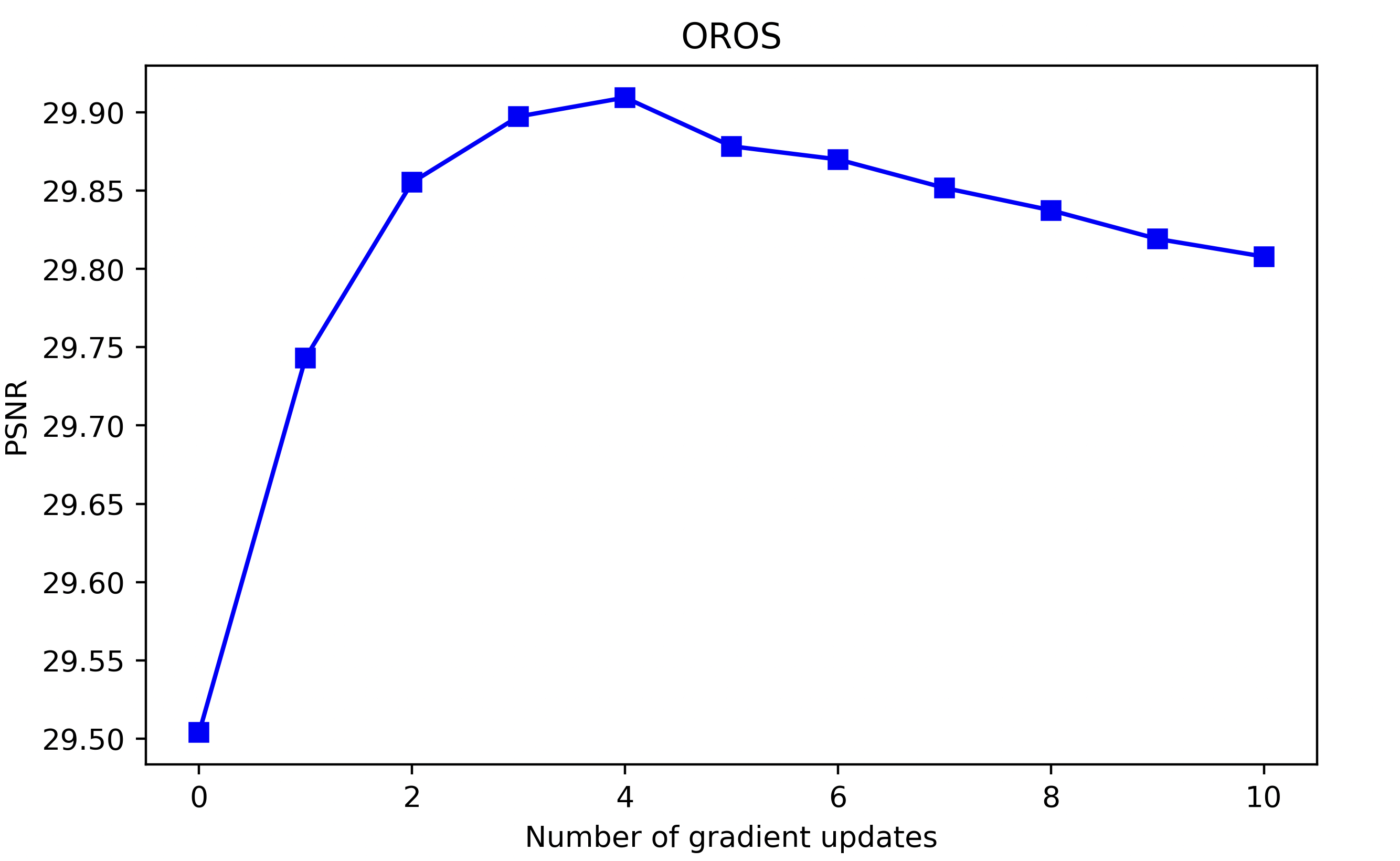}
    \includegraphics[width =0.45\linewidth]{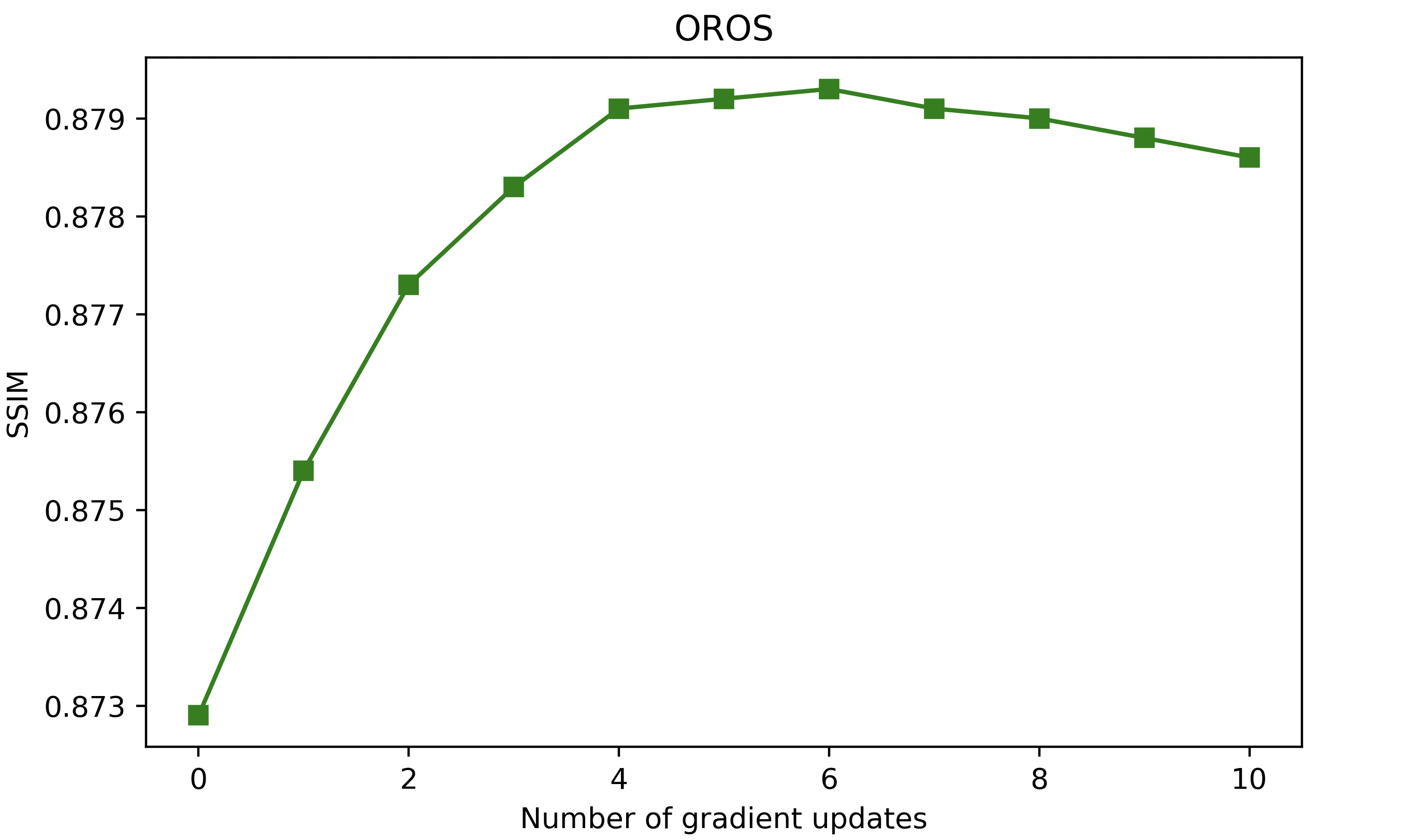}
    \caption{\textbf{Illustration of PSNR and SSIM after each gradient update.} Our method can utilize more frames in the testing time for better performance until convergence. Note that we get the best performance when gradient update times $\mathbf{n}=4$.} 
    \label{fig:test time updates}
\end{figure}

\noindent\textbf{Number of Gradient Updates}: Here, we study the impact of the number of gradient updates on testing time.   
The results of the qualitative comparison of different test time updates on four sub-datasets of OROS are shown in Fig.~\ref{fig: n-update}.
The model is trained with various numbers of gradient updates ($\mathbf{n}$ from 1 to 10). 
According to the results, it seems that larger $\mathbf{n}$ tend to get better results and contribute to adapting to the occlusion features of a test image.
Then, we illustrate the quantitative evaluation after each gradient update, as shown in Fig.~\ref{fig:test time updates}.
Contrarily, large $\mathbf{n}$ will result in the model overfitting during test time adaptation in Fig.~\ref{fig:test time updates}.
It can be found that the reasonable times of gradient update for our situation $\mathbf{n}$ is 4.

\section{Conclusion}
In this paper, we first formulate a physical model for occlusions removal through unclear glass. We then proposed a neural network with two branches that shares the same parameters of an auto-encoder. By updating via an auxiliary branch during test phase, the model is able to perform test-time  adaptation for enhanced occlusion removal. To further facilitate effective supervised learning and robust evaluation, we developed OROS, a realistic ground-truth glass obstacle dataset. Extensive experiments indicate  that the suggested method outperforms the state-of-the-art methods quantitatively and qualitatively on removing realistic occlusions, especially the unseen ones.

{\small
\bibliographystyle{ieee_fullname}
\bibliography{main}
}

\end{document}